\documentclass[11pt]{article}

\usepackage[final]{acl}

\usepackage{times}
\usepackage{latexsym}
\usepackage{amsmath}
\usepackage{wrapfig}
\usepackage{booktabs}
\usepackage{caption}

\newcommand{\appref}[1]{Supplementary \ref{#1}}
\newcommand{\secref}[1]{Section \ref{#1}}
\newcommand{\tabref}[1]{Table \ref{#1}}
\newcommand{\figref}[1]{Figure \ref{#1}}

\newcommand{\myex}{{\em The Hermit of Far End}}

\usepackage[T1]{fontenc}
\usepackage[utf8]{inputenc}

\usepackage{microtype}

\usepackage{inconsolata}

\usepackage{graphicx}

\title{Repeated Sequences Reveal Gaps \\
  between Large Language Models and Natural Language}

\author{Kumiko Tanaka-Ishii \\
  Waseda University, Japan. \\
  \texttt{kumiko@waseda.jp}}

\begin{document}
\maketitle

\begin{abstract}
Evaluating whether large language models (LLMs) capture the structure
of natural language beyond local fluency remains an open challenge.
Existing evaluation methods, largely based on task performance or
short-context behavior, provide limited insight into the long-range
statistical organization of generated text.

We propose a complementary evaluation framework based on repeated
subsequences. By analyzing their distribution across scales and
relating it to higher-order R\'enyi entropies, we probe how texts
reuse previously established structure under finite-length
conditions. Experiments on human-written texts and length-matched
GPT-generated texts show that,
while power-law models can describe
restricted ranges of block length, the observed entropy growth is
often equally or better characterized by logarithmic--power forms.

Across datasets, natural language exhibits stable entropy-growth
patterns over accessible ranges, with consistent average 
behavior despite variability across individual texts. In contrast,
GPT-generated texts show systematic and statistically significant
shifts in estimated exponents with model size.
These results demonstrate that repeated-subsequence entropy
provides a quantitative structural diagnostic that reveals
systematic differences in long-range organization,
distinguishing natural language from state-of-the-art LLM outputs
beyond surface-level fluency.
\end{abstract}

\section{Introduction}
\label{sec:introduction}

Recent large language models (LLMs) generate highly fluent and
coherent text, achieving strong performance across a wide range of
language tasks. However, it remains unclear whether such models
capture the long-range organization of natural language beyond local
consistency. Improvements in next-token prediction do not necessarily
imply that generated texts exhibit human-like structure at larger
scales.

Most evaluations of language models rely on task-based benchmarks or
short-context analyses, which primarily measure task performance
\cite{Brown2020,Bubeck2023}. While effective for downstream
performance, these approaches provide only indirect evidence about the
global statistical structure of generated text.  Prior work has
identified systematic issues in generated text, including excessive
repetition and reduced diversity \cite{Holtzman2019,Welleck2020},
suggesting that high task performance does not necessarily imply
human-like long-range organization.


These limitations suggest that, while LLMs are effective at producing
locally coherent outputs, they may struggle to sustain globally
consistent organization over long spans.  In natural language,
expressions are not used in isolation but are repeatedly referred to,
reused, and recombined in different contexts. Such organization can be
understood as a form of \emph{reference structure}, in which
previously established elements are revisited and integrated over long
distances.

Since reference structure is primarily realized through reuse, it
should be observable as repetition in the sequence. We therefore
analyze repetition as a distributional property across scales,
capturing how subsequences of different lengths are reused throughout
a text.

Repetition has long been recognized as a fundamental property of
symbolic sequences in information theory
\cite{ZivLempel1978,OrnsteinWeiss1993}. Building on this
line of work, we relate repetition statistics to higher-order Rényi
entropies, providing a finite-length characterization of entropy
growth. As discussed in Section~\ref{sec:theory}, different growth
regimes correspond to qualitatively distinct types of structural
organization.

Using this framework, we compare natural language and LLM-generated
texts. We find that natural language exhibits stable entropy growth,
whereas LLM-generated text shows systematic shifts in 
exponents with model size, indicating differences in how structure is
reused over long ranges.

Our contributions are as follows:
(1) we propose a distributional formulation of repetition that
enables a finite-length characterization of entropy growth via
higher-order Rényi entropies; and
(2) we demonstrate systematic differences between natural language
and LLM-generated text in entropy growth behavior, revealing
distinct patterns of structural reuse.

\section{Related Work}
\label{sec:related}

\subsection{Evaluating long-range structure in language models}

Most evaluations of language models rely on task-based benchmarks or
short-context analyses, which primarily measure task performance
\cite{Brown2020,Bubeck2023}. While effective for assessing downstream
performance, these approaches provide only indirect evidence about the
global statistical organization of generated text.

A complementary line of work has examined language models from an
information-theoretic perspective, analyzing properties such as
entropy rates, calibration, and memory usage \cite{Braverman2020}, as
well as inductive biases toward long-range dependencies
\cite{Hahn2020,Merrill2021}.  Recent studies have also tackled
limitations in effectively utilizing long contexts and maintaining
coherent structure across extended sequences \cite{Press2022}.

While these approaches provide valuable insights into model behavior,
they do not directly address how generated texts reuse structure across
scales. Our work addresses this gap by focusing on repeated
subsequences as a distributional signal of structural reuse in long
texts.

\subsection{Repetition in natural and generated texts}

Repetition has long been studied as a signature of long-range
structure in symbolic sequences. Related ideas arise in universal
compression schemes such as Lempel--Ziv coding, where repeated
subsequences determine compression performance
\cite{ZivLempel1978}.

Analyses based on maximal repeated subsequences capture extreme
instances of such repetition.  For i.i.d. sources over a finite
alphabet, maximal repetition grows logarithmically with sequence
length \cite{OrnsteinWeiss1993,WynerZiv1989}, whereas natural language
exhibits stronger maximal repetition growth
\cite{DebowskiMR}.  However, such statistics are numerically unstable
in finite texts, which limits their use as a primary diagnostic.

In the context of neural text generation, repetition has also been
studied as a generation artifact, including degeneration and
excessive repetition \cite{Holtzman2019,Welleck2020}. Subsequent work
has further shown that such behaviors are strongly influenced by the
choice of decoding strategy, which can lead to qualitatively different
trade-offs between diversity, coherence, and repetition across tasks
\cite{Wiher2022}. These approaches focus on mitigating or
regulating repetition, rather than analyzing it as a structural
property of the sequence.

Overall, while repetition has been examined through extreme
statistics, compression-based methods, and generation behavior,
systematic distributional analyses of repeated subsequences across
scales remain limited. This gap motivates the present work.

\subsection{Entropy scaling in natural language}

A complementary perspective on language structure is to examine how
information grows as the context length increases.

A classical approach to long-range structure analyzes the scaling of
block entropy, typically using Shannon entropy
\cite{Shannon1948,Shannon1951}.
Empirical studies report sublinear entropy growth in natural language,
often discussed in the context of power-law-like behavior
\cite{Hilberg1990,DebowskiEntropyRate}.
Entropy
estimates converge extremely slowly
\cite{DebowskiEntropyRate,TakahiraTanakaIshiiDebowski2016}, leaving open
whether observed scaling reflects asymptotic behavior or finite-size
effects.

Higher-order Rényi entropies have also been considered in the
theoretical analysis of entropy rates \cite{DebowskiEntropyRate}, and in
empirical studies of linguistic statistics. In particular,
\citet{TanakaIshii2015} show that certain Rényi-based measures, such
as Yule’s K, exhibit approximate constancy with respect to text
length, highlighting scale-invariant properties of lexical
distributions.

However, prior work has primarily used Rényi entropies either in
theoretical settings or as static measures of distributions, rather
than to characterize how entropy varies with context length.
In contrast, the present work uses Rényi entropies to analyze
entropy growth as a function of context length, linking them to the
distribution of repeated subsequences and enabling a finite-length
characterization of scaling behavior.

\section{Three Entropy Growth Regimes and Structural Reuse}
\label{sec:theory}

This paper studies how information grows as longer contexts are
considered. Let $\mathcal{A}$ be a finite alphabet, and let
$X = x_1, x_2, \dots, x_n$ be a sequence of length $n$ over
$\mathcal{A}$. We call a contiguous subsequence of length $m$ a
\emph{block}. Let $H_1(m)$ denote the Shannon entropy of blocks of
length $m$. The growth rate of $H_1(m)$ characterizes how much new
information is introduced as the block length increases.

Entropy growth is commonly decomposed into an extensive linear term
and a subextensive correction \cite{Hilberg1990,DebowskiBook}
\[
H_1(m) = h_1 m + G(m), \label{eq:h}
\]
where $h_1$ is the entropy rate and $G(m)$ captures deviations from
linear growth. If the subextensive term satisfies $G(m) = o(m)$,
then $H_1(m)/m \to h_1$ as $m \to \infty$.
It remains an open question whether the entropy rate $h_1$ of
natural language is strictly positive; in particular, it has been
conjectured that it may vanish under Hilberg-type scaling
\cite{Hilberg1990,DebowskiBook}.
For finite and empirically accessible ranges of $m$, the subextensive
term $G(m)$ may dominate the observed behavior and thus captures the
effective structure of entropy growth.

Three qualitatively distinct regimes have been considered for
$G(m)$. First, for i.i.d.\ or finite-order Markov systems,
$G(m) = O(1)$. Second, systems with long-range dependencies and
expanding structural degrees of freedom exhibit sublinear growth,
often approximated by a power law \cite{Hilberg1990}:
\begin{equation}
G(m) \propto m^{\beta}.  \label{eq:power}
\end{equation}
This regime has been linked to grammar-based structure in natural
language, where hierarchical rules and expanding sets of patterns give
rise to power-law growth \cite{DebowskiBook}.

Third, studies of predictive information and complexity have
identified logarithmic or log-power growth as a distinct regime,
\begin{equation}
G(m) \propto (\log m)^{\gamma}, \label{eq:logpower}
\end{equation}
associated with strong structural reuse \cite{Bialek2001}. Such growth
is consistent with systems whose effective description length
increases much more slowly than sequence length
\cite{Kolmogorov1965,LiVitanyi2008}.  While power-law behavior has
been discussed for natural language, as mentioned previously,
logarithmic or log-power scaling has
not been systematically explored in empirical studies.

These regimes are formulated for Shannon entropy but are speculated to
extend to higher-order entropy measures, which are derived from the
same block distribution. Differences between entropy orders primarily
reflect different weighting of frequent versus rare patterns, while
the overall regime is governed by the growth of distinct blocks with
length.

In this work, we show that higher-order entropy measures are
compatible with a logarithmic-power form, suggesting that entropy
growth in natural language lies near this regime at accessible
sequence lengths. As discussed in Section~\ref{sec:discussion}, this
behavior admits an interpretation in terms of strong structural reuse,
where texts are generated through recombination and re-indexing of
shared linguistic resources.

\section{Proposed Method}

\subsection{Counting Repeated Subsequences}
\label{sec:binaryDup}

We formally define the number of repeated subsequences and
propose a method to characterize their behavior, which reveals
systematic differences between natural language texts and
state-of-the-art GPT-generated texts.

As mentioned previously, let $X = x_1, x_2, \ldots, x_n$
be a sequence over $\mathcal{A}$, a finite set of alphabet. 
For $1 \leq i < j \leq n$, let
$x_i^j$ denote the contiguous subsequence from position $i$ to
$j-1$. A subsequence is said to be repeated if two subsequences of
length $m$ starting at distinct positions $i$ and $j$ coincide, i.e.,
$x_i^{i+m} = x_j^{j+m}$ for some $i \neq j$.

Let $m$ denote the length of a repeated subsequence, and call any
consecutive subsequence a \emph{block}. In a sequence of total length
$n$, there are $T_m = n - m + 1$ blocks of length $m$. Let $K_m$ denote
the number of \emph{distinct} block types of length $m$. The number of
repetitions of length-$m$ blocks is then defined as
\begin{equation}
  D_m = T_m - K_m . \label{eq:dtkm}
\end{equation}

For example, for a sequence $X=\text{``banana''}$ of length $n=6$, when
$m=2$ the blocks are $\text{ba}, \text{an}, \text{na}, \text{an},
\text{na}$, yielding $T_2 = 5 = 6 - 2 + 1$. There are three distinct
blocks, $\text{ba}, \text{an}, \text{na}$, so that $K_2 = 3$ and
$D_2 = 2$. Indeed, the blocks ``an'' and ``na'' are each repeated once;
their total number of repetitions is $D_2$ for the sequence
``banana''.
Since $1 \leq K_m \leq T_m$, where $K_m = 1$ corresponds to the case in
which all blocks are identical and $K_m = T_m$ corresponds to the case
in which all blocks are distinct, the range of $D_m$ is
$0 \leq D_m \leq T_m - 1$.

\begin{figure}[t]
  \centering
  \vspace*{-5mm}
  \includegraphics[width=0.9\columnwidth]{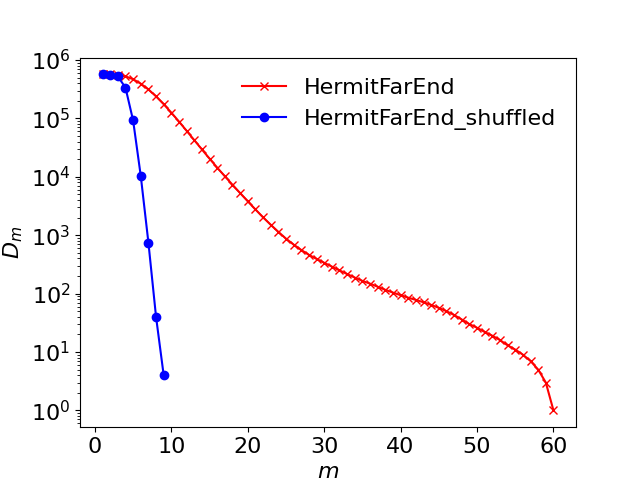}
  \vspace*{-3mm}
  \caption{Number of repeated blocks $D_m$ as a function of block
    length $m$ for \myex\ (red) and its randomly shuffled counterpart
    (blue). \label{fig:hfdDup}}
\end{figure}

\figref{fig:hfdDup} shows the observed values of $D_m$ (vertical axis,
logarithmic scale) as a function of $m$ (horizontal axis) for
\myex\ (by Margaret Pedler, $n=586{,}533$, red) and its shuffled
version (blue). The shuffled text exhibits behavior similar to that of
a Bernoulli process, as shown in \appref{app:binaryDup}. For small
values of $m$, essentially all $|\mathcal{A}|^m$ possible blocks
appear in the sequence. As $m$ increases, the number of repetitions
decreases sharply.


In contrast, natural language displays a markedly different and
strongly non-linear pattern. In particular, repetitions persist up to
block lengths close to $60$, indicating that while the maximally
repeated subsequence can be very long, the distribution of shorter
repeated blocks already reflects characteristic structural properties
of the text.

Nevertheless, the absolute vertical position of $D_m$ is dominated by
the document length $n$, and its non-linear shape is difficult to
capture with a simple functional form. Since the vertical axis is
plotted on a logarithmic scale, this observation naturally motivates
an information-theoretic characterization of repetition, which we
develop in the next subsection.

\subsection{Higher-order R\'enyi Entropy}
\label{sec:renyi}

Let $S_m$ denote the total number of possible block types of length $m$
over the alphabet $\mathcal{A}$. For a completely random sequence,
$S_m = |\mathcal{A}|^m$, whereas for structured systems such as natural
language, typically $S_m \ll |\mathcal{A}|^m$. Among these $S_m$
possible blocks, only a subset actually appears in a given sequence;
this observed subset is counted by $K_m$. Let $p_w$ denote the
probability of occurrence of a particular block $w$. For the moment,
we restrict attention to blocks of fixed length $|w| = m$. The
probability that $w$ does not appear in any of the $T_m$ positions is
$(1-p_w)^{T_m}$, and hence the probability that it appears at least
once is $1-(1-p_w)^{T_m}$, whose sum  over $|w|=m$ is $E[K_m]$,
the expected number of distinct observed blocks. 
Therefore, from formula~(\ref{eq:dtkm}),
\begin{equation}
  E[D_m] = T_m - \sum_{|w|=m} \left(1-(1-p_w)^{T_m}\right).
  \label{eq:dm}
\end{equation}

For a sufficiently long sequence, $T_m$ is large, and thus
$(1-p_{w})^{T_m} \approx e^{-T_m p_{w}}$, which can be expanded as
\begin{equation*}
  e^{-T_m p_{w}} = 1 - T_m p_{w} + \frac{T_m^2 p_{w}^2}{2!}
  - \frac{T_m^3 p_{w}^3}{3!} + \cdots .
\end{equation*}
Substituting this expansion into~(\ref{eq:dm}), the first-order terms
cancel, yielding
\begin{equation}
E[D_m] \approx \sum_{|w|=m}\left(
  \frac{T_m^2 p_{w}^2}{2!}
  - \frac{T_m^3 p_{w}^3}{3!} + \cdots
\right).
\label{eq:expand}
\end{equation}
Therefore, $E[D_m]$ is characterized by the spectrum
$\sum_{|w|=m} p_w^{\alpha}$ with $\alpha \geq 2$.

As noted above, we wanted to analyze $E[D_m]$ on a logarithmic scale.
This spectrum is naturally captured by the R\'enyi entropy
of order $\alpha$~\cite{Renyi1961}, defined as
\begin{equation}
  H_{\alpha}(m) = \frac{1}{1-\alpha}
  \log_2 \sum_{|w|=m} p_w^{\alpha}.
\end{equation}
Larger values of $\alpha$ place greater weight on frequently occurring
blocks, whereas smaller values emphasize overall diversity.  As is
well known, $H_{\alpha}(m) \to H_1(m)$ as $\alpha \to 1$.
Furthermore, if the underlying stochastic process is stationary and
ergodic, and if the Rényi entropy rate $h_{\alpha}$ exists,
then $H_{\alpha}(m)/m \to h_{\alpha}$ as
$m \to \infty$ \cite{CoverThomas2006}.
For a Bernoulli process with $p=0.5$, we have
$h_{\alpha} = 1$ for all $\alpha$ (see \appref{app:halpha1}).

\begin{figure}[t]
  \centering
  \includegraphics[width=0.96\columnwidth]{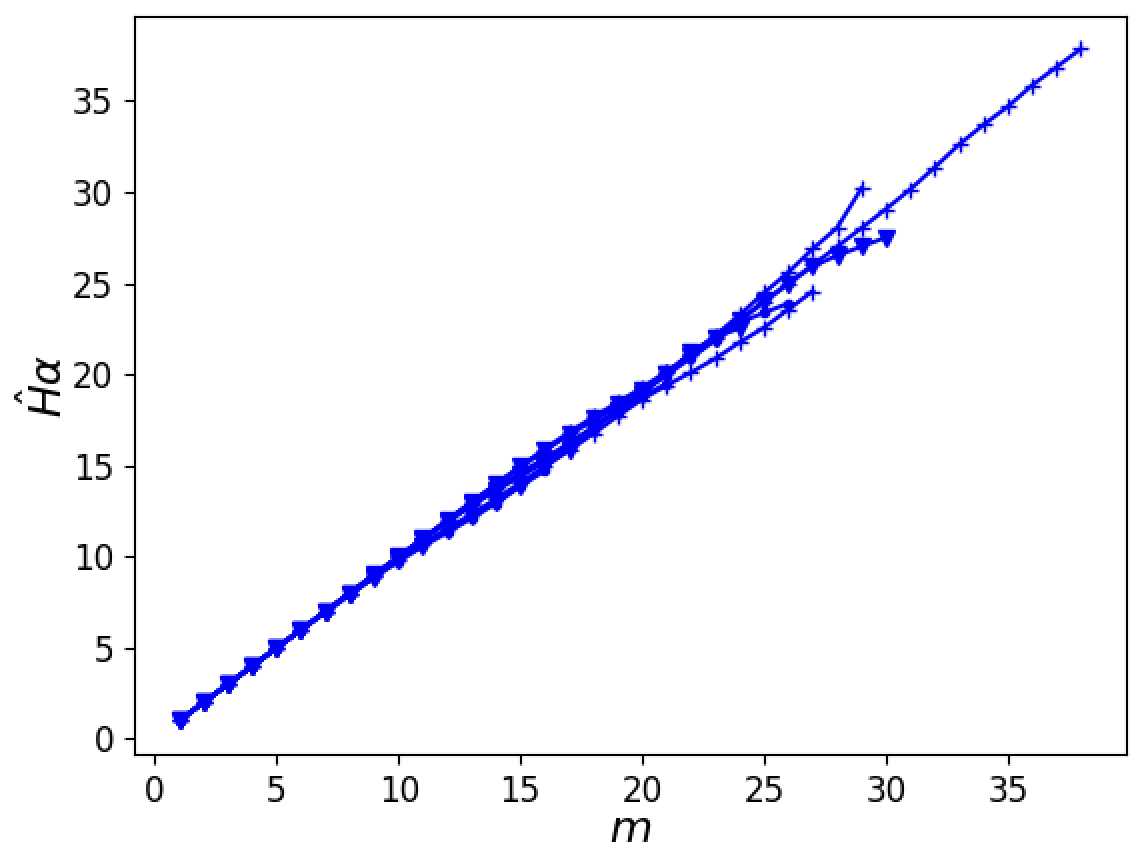}
  \includegraphics[width=0.96\columnwidth]{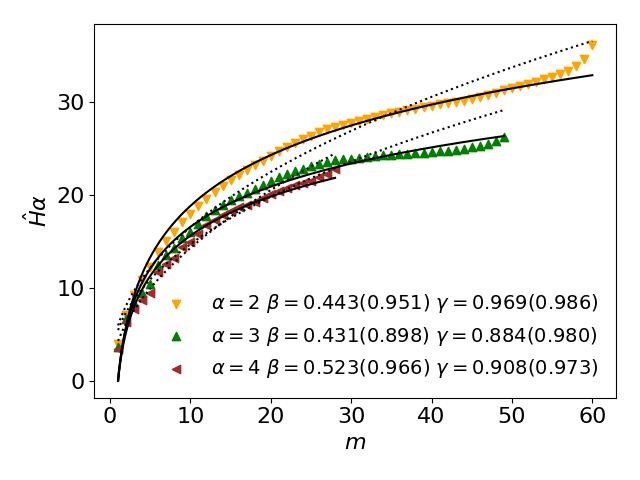}
  \vspace*{-5mm}
  \caption{Empirical higher-order R\'enyi entropy
    $\hat{H}_{\alpha}(m)$ for $\alpha = 2,3,4$.
    Top: Bernoulli process with $p=0.5$ for sequence lengths
    $10\mathrm{k}$, $100\mathrm{k}$, and $1\mathrm{M}$ (all curves
    collapse). Bottom: \myex.
    \label{fig:h2-hfe}}
  \vspace*{-3mm}
\end{figure}

\figref{fig:h2-hfe} shows the empirical spectra
$\hat{H}_{\alpha}(m) = \frac{1}{1-\alpha}
\log_2 \sum_{i=1}^{K_m} \hat{p}_{w_i}^{\alpha}$, where
$\hat{p}_{w} \equiv \#w / T_m$ and $\#w$ denotes the frequency of block
type $w$ in the sequence, restricted to $\#w \geq \alpha$. The
rationale for approximating $p_w$ by this empirical distribution is
deferred to \appref{app:p}, and we proceed with the main text.

The top panel shows higher-order R\'enyi entropies for three Bernoulli
processes of different lengths, with $p=0.5$ and $\alpha=2,3,4$,
yielding nine curves in total. All curves collapse onto a single line
with slope~1.  In contrast, the bottom panel shows
$\hat{H}_{\alpha}(m)$ for \myex, where $\alpha=2,3,4$ are shown in
yellow, green, and brown, respectively. The shuffled versions of
\myex\ exhibit linear collapse similar to the Bernoulli case and are
omitted for clarity. Larger values of $\alpha$ yield shorter curves
due to the restriction $\#w \geq \alpha$.

\subsection{Functional Characterization of Spectrum}

We now seek a functional description of the empirical spectrum
$\hat{H}_{\alpha}(m)$.  The empirical $\hat{H}_{\alpha}(m)$ increases
smoothly over a wide range of $m$, but shows a sharp rise as it
approaches the upper bound imposed by the total number of blocks
$T_m$.

Following the derivation in \appref{app:delta}, it can be shown that
\begin{equation}
  H_{\alpha}(m) \approx \log_2 S_m - \Delta_{\alpha},
  \label{eq:base}
\end{equation}
where $\Delta_{\alpha}$ is expressed as a weighted logarithm of
Touchard polynomials. For example, for $\alpha=2$ we have $\Delta_2 =
\log_2 \left( 1 + \frac{1}{\lambda_m} \right)$ with $\lambda_m \equiv
T_m/S_m$, while $\Delta_{3}$ and $\Delta_{4}$ are given in the same
appendix. The term $\Delta_{\alpha}$ therefore represents a
finite-size correction arising from the upper bound imposed by
$T_m$. Consequently, we focus on modeling the leading term $\log_2
S_m$, which captures the effective growth of the the number of
distinguishable blocks underlying entropy scaling.

The classification in \secref{sec:theory}, given by
(\ref{eq:power}) and (\ref{eq:logpower}), suggests two distinct forms
of sublinear entropy growth.  The first model is the power-law ansatz:
\begin{equation}
  \log_2 S_m \propto m^{\beta},
  \label{eq:model1}
\end{equation}
where $\beta$ is the exponent. This form corresponds to systems in
which the effective number of distinguishable blocks continues to
expand with block length, reflecting increasing structural degrees
of freedom.
The second model is the logarithmic--power ansatz:
\begin{equation}
  \log_2 S_m \propto (\log_2 m)^{\gamma},
  \label{eq:model2}
\end{equation}
where $\gamma$ is the exponent. This form corresponds to entropy
growth dominated by the reuse and recombination of previously
established structure.

We do not include an explicit linear term $h m$ in the regression (see
\secref{sec:theory}), because over the finite range of $m$ considered,
it is not reliably separable from the subextensive component.  As our
focus is on entropy growth behavior, we model $\log S_m$ directly
using sublinear forms.


Direct estimation of $\beta$ or $\gamma$ from $K_m$ alone is not
feasible, since $K_m \ll S_m$ for large $m$.  Alternatively,
estimating these exponents directly from (\ref{eq:base}) by
substituting the ansatz leads to unstable results, because the
logarithmic correction term $\Delta_{\alpha}$ can be non-negligible at
finite lengths.  We therefore adopt a two-stage estimation procedure:
\begin{enumerate}
\item Estimate $\lambda_m$ from its functional relation with
  $D_m/T_m$, as derived in \appref{app:lambda}, and
\item Estimate $\beta$ and $\gamma$ by fitting $\log_2 S_m$ in
  $\hat{H}_{\alpha}(m) = \log_2 S_m - \Delta_{\alpha}$, where
  $\Delta_{\alpha}$ depends only on $\lambda_m$ and
  $\hat{H}_{\alpha}(m)$ is empirically estimated.
\end{enumerate}

\figref{fig:h2-hfe} (bottom) shows the resulting fits to
$\hat{H}_{\alpha}(m)$ for $\alpha=2,3,4$ for \myex. Power-law fits are
shown as dotted lines, while log-power fits are shown as solid lines.
The power-law model systematically overestimates the growth rate,
particularly for $\alpha=2$ and $3$, where it increases too rapidly at
large $m$. In contrast, the log-power model captures the overall trend
substantially better across all values of $\alpha$. The legend reports
the estimated values of $\beta$ and $\gamma$, together with the
coefficient of determination, as follows:
\[
R^2 = 1 - \frac{\sum_{i} (y_i - \hat{y}_i)^2}
               {\sum_{i} (y_i - \bar{y})^2}, 
               \]
where, $y_i$ are observed values
($\hat{H}_{\alpha}(m)+\Delta_{\alpha}$), $\hat{y}_i$ are
fitted values to the model (power or log-power of $m$), and $\bar{y}$
is the mean of the observed values.  Values closer to $1$ indicate a
better fit.  For \myex, although the power-law model (dotted lines)
yields $R^2$ values in the range $0.90$--$0.96$, its deviations are
visually larger than those of the log-power model (solid lines), for
which $R^2 > 0.97$.  Hence, the log-power model is preferred over the
power-law model for all tested values of $\alpha$.

As we will show below, empirical estimates of $\hat{H}_{\alpha}(m)$
for natural language often lie near the boundary between these two
regimes, making it difficult to distinguish between a pure power-law
and a log-power form at accessible sequence lengths.

\section{Experiments}

We evaluate long-range organization in text by analyzing
repeated-subsequence statistics.  Differences between natural language
and LLM outputs are assessed via fitted entropy-growth parameters and
statistical tests.  All comparisons are conducted on collections of
long texts with matched length distributions.

\begin{table}[t]
\centering
  \caption{Datasets with the mean and standard deviation of text
    length. 
    Natural language texts are sampled from Project
    Gutenberg to match the length distribution of the corresponding
    GPT datasets.}
  \label{tab:dataset}
  \vspace*{-2mm}
  \footnotesize
  \begin{tabular}{|l|l|r|}
    \hline
  dataset & number  & length (chars) \\
    \hline
    \multicolumn{3}{|c|}{GPT-generated text} \\
    \hline
    gpt-3.5turbo & 100 & 35044.91$\pm$2287.31\hspace*{1.75mm}  \\
    gpt-4o-mini  & 100 & 110888.61$\pm$23378.64 \\
    gpt-5-mini   & 100 & 347044.85$\pm$19793.48 \\
    gpt-5        & 100 & 601187.13$\pm$24972.87 \\
    \hline
    \multicolumn{3}{|c|}{Natural language text} \\
    \hline
    nl-3.5turbo  & 100 & 34741.86$\pm$1997.81\hspace*{1.75mm}  \\
    nl-4o-mini   & 100 & 108762.77$\pm$24223.76 \\
    nl-5-mini    & 100 & 346913.54$\pm$18034.12 \\
    nl-5         & 100 & 593630.96$\pm$27223.67 \\
    \hline
  \end{tabular}
\end{table}

\subsection{Dataset}
\label{sec:dataset}

Our analysis targets entropy growth over a wide range of block
lengths, which requires long and coherent sequences. This
substantially restricts the choice of data: among naturally occurring
texts, extended narratives such as novels provide one of the few
sources of sufficiently long sequences.  For the same reason, previous
studies of maximal repetition have also focused on novels
\cite{DebowskiMR}, which provides an additional motivation for our
choice of data.


Based on these considerations, we analyze both natural language texts
and texts generated by GPT models. Specifically, we consider outputs
from gpt-3.5turbo, gpt-4o-mini, gpt-5-mini, and gpt-5, as well as
human-written natural language texts. Earlier models such as gpt-1 and
gpt-2 are not included, as they are unable to reliably generate texts
of sufficient length.

For non-stationary data, the behavior of $H_{\alpha}(m)$ may still
depend on the overall sequence length $n$. However, as we will show
below, the estimated exponents for natural language exhibit a
degree of universality. To control for length effects, we first
sampled long stories from each GPT model under the generation settings
described in \appref{app:gpt}. For comparison, we sampled natural
language texts from the Project Gutenberg corpus, so that length
distributions are matched to those of corresponding gpt-X.

Before sampling, we preprocessed all texts from the Standardized
Project Gutenberg Corpus \cite{GerlachFontClos2018SPGC} by removing
metadata and layout-related whitespace. We restrict our analysis to
languages that use alphabetic writing systems, leaving extensions to
non-alphabetic scripts for future work. For each data category, we
sampled 100 texts.

Finally, we note a potential domain-related consideration arising from
the use of narrative texts.  As mentioned, long coherent texts are not
abundantly available outside of narrative genres, which is the main
reason for our choice.  To assess whether our findings are specific to
narratives, we conducted additional experiments using GPT-generated
texts under alternative prompts, including essays and scientific-style
writing.  These results are reported in \appref{app:domain}.  Although
such texts are typically much shorter, the overall trends
remain almost consistent with those observed for narratives.

\subsection{$\beta$ and $\gamma$}
\label{sec:betagamma}

\begin{figure}[t]
  \centering
  \includegraphics[width=\columnwidth]{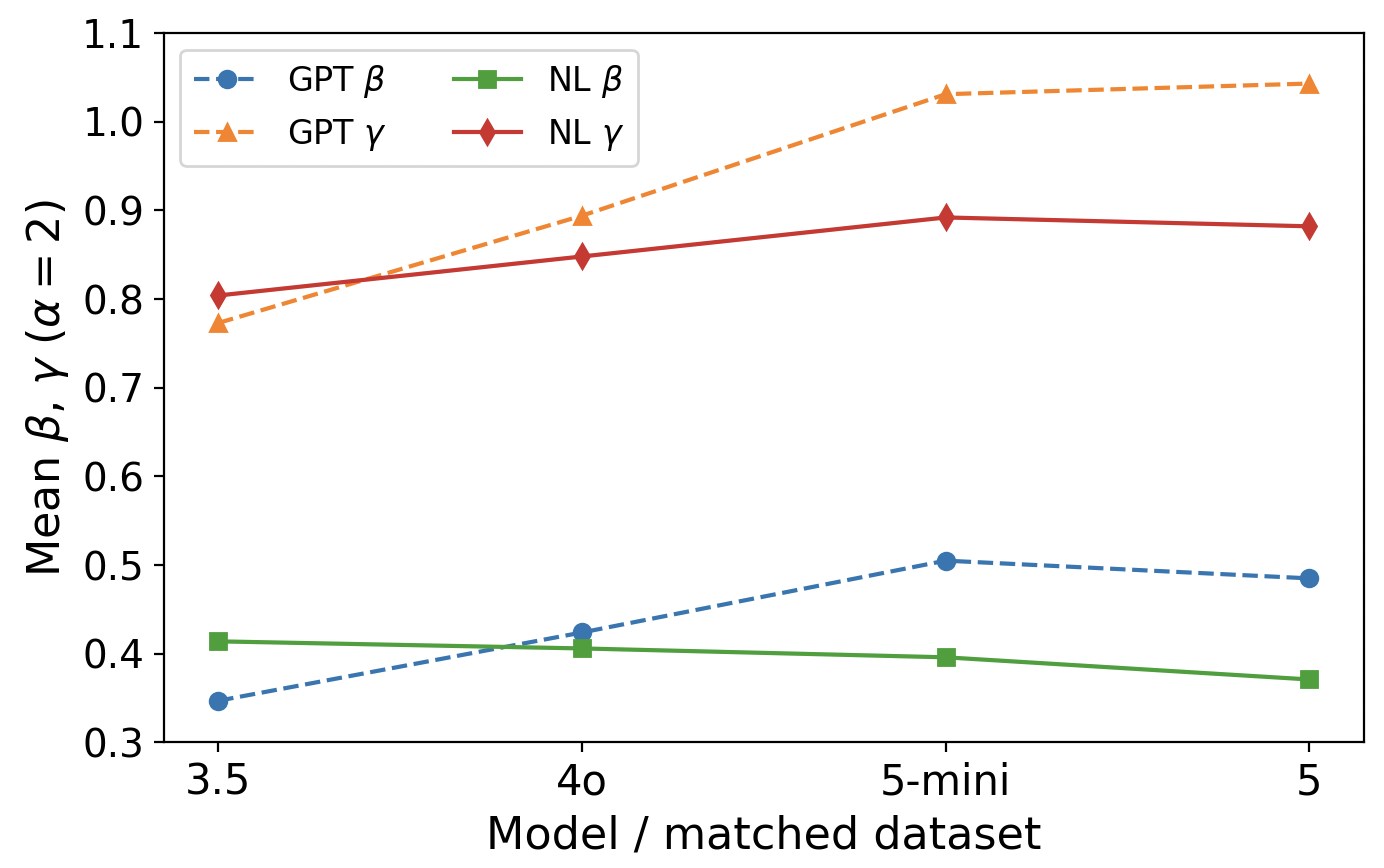}
  \vspace*{-6mm}
  \caption{Mean exponents $\beta$ and $\gamma$ for $\alpha=2$.
  Natural language remains approximately stable across datasets,
  whereas GPT-generated text exhibits a monotonic increase with model
  size. This figure summarizes the central tendency of the
  distributions shown in Figures~\ref{fig:boxbeta} and
  \ref{fig:boxgamma}.}
  \label{fig:summary}
  \vspace*{-0.3cm}
\end{figure}

\begin{figure}[t]
  \centering
  \includegraphics[width=\columnwidth]{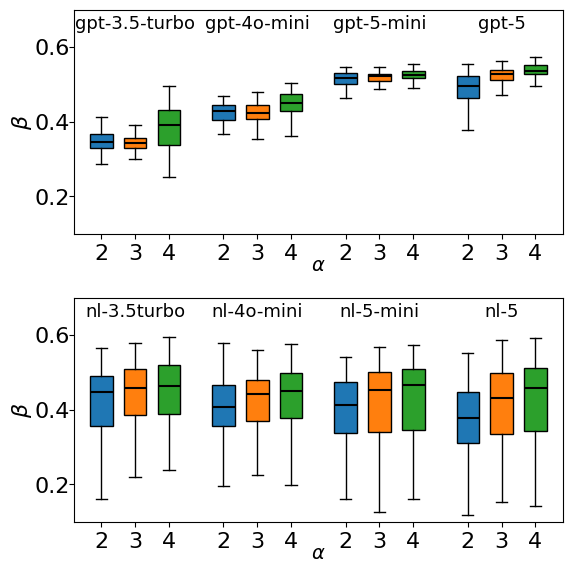}
\vspace*{-8mm}
  \caption{Boxplots of $\beta$ for each dataset.
    \label{fig:boxbeta}}
\vspace*{-0.5cm}
\end{figure}

\begin{figure}[t]
  \centering
  \includegraphics[width=\columnwidth]{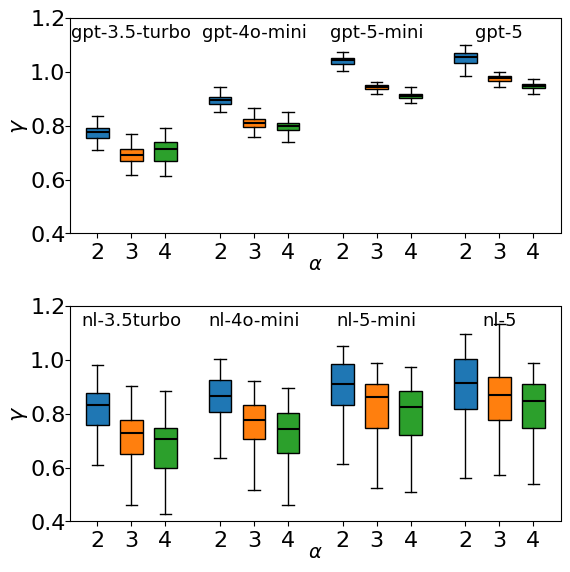}
\vspace*{-8mm}
  \caption{Boxplots of $\gamma$ for each dataset.
    \label{fig:boxgamma}}
\vspace*{-3mm}
\end{figure}

Figure~\ref{fig:summary} provides a consolidated summary view of the mean
exponents $\beta$ and $\gamma$ for $\alpha=2$. Natural
language remains approximately stable across datasets despite large
differences in text length, whereas GPT-generated text exhibits a
clear monotonic increase in both $\beta$ and $\gamma$ with model size.
For gpt-5 and gpt-5-mini, both $\beta$ and $\gamma$ are substantially
larger than for natural language.

This contrast is further supported by the full distributions shown in
\figref{fig:boxbeta} and \figref{fig:boxgamma} (See \appref{app:vals}
for numerical values).  Natural language displays substantial
variability across individual texts but maintains stable mean values,
suggesting a form of weak universality in which growth behavior
emerges at the population level.  In contrast, GPT-generated texts are
more homogeneous and exhibit a systematic shift toward larger exponent
values as model size increases from gpt-3.5 to gpt-5.  As a reference,
representative examples of $\hat{H}_{\alpha}(m)$ for individual
GPT-generated text samples are shown in \appref{app:gpthalpha}. These
examples provide qualitative context for interpreting the large-scale
statistical analyses in this section.

Welch two-sample $t$-tests indicate clear statistical separation
between GPT-generated and natural language texts (gpt-5 vs.\ nl-5 and
gpt-5-mini vs.\ nl-5-mini, with $p \approx 0$ for both $\beta$ and
$\gamma$).  Within natural language, comparisons
between nl-5 vs.\ nl-5-mini do not yield statistically
significant differences ($p=0.12$ for $\beta$, $p=0.94$ for $\gamma$).
This is consistent with higher variability but no systematic shift in
exponent values, in contrast to the monotonic increase observed for
GPT models.

Turning to the dependence on the R\'enyi order $\alpha$, we observe a
systematic increase of $\beta$ with $\alpha$. Since larger $\alpha$
places greater weight on high-probability events, this indicates that
frequent blocks exhibit stronger effective growth in structural
degrees of freedom.  In contrast, the parameter $\gamma$ decreases
with $\alpha$, implying that lower-order entropies, which remain
sensitive to rare events, exhibit more pronounced convexity. This
suggests that rare patterns are more strongly governed by structural
reuse, leading to slower effective growth.  Taken together, these
trends indicate that entropy growth is not uniform across the
distribution of patterns: frequent and rare structures follow
qualitatively different behaviors.
This heterogeneity is observed in both natural language and
GPT-generated text.

\begin{figure}[t]
  \centering
  \includegraphics[width=\columnwidth]{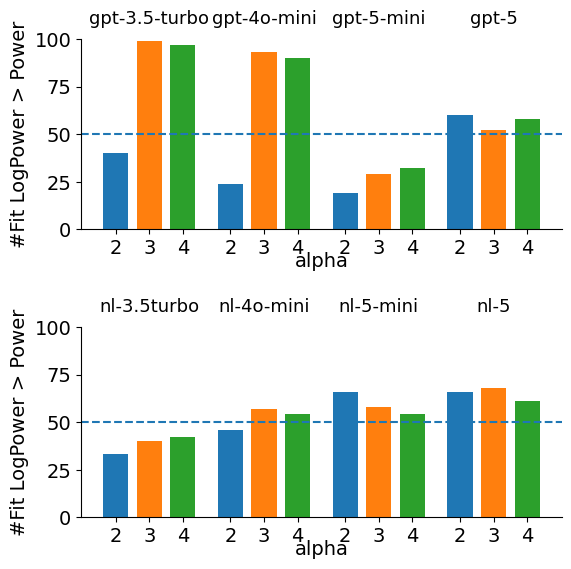}
  \vspace*{-8mm}
  \caption{Number of texts for which the log-power model achieves a
    higher coefficient of determination than the power-law model.}
  \label{fig:fit}
  \vspace*{-3mm}
\end{figure}

Across all datasets, the average coefficient of determination $R^2$
exceeds $0.87$ and is typically close to $0.95$. \figref{fig:fit}
summarizes model preference by counting the number of texts for which
the log-power fit achieves a higher $R^2$ than the power-law fit (see
\appref{app:vals} for numerical values).

For GPT-generated texts (\figref{fig:fit} top), smaller
models show a strong preference for the log-power model at $\alpha=3$
and $4$, suggesting the presence of many repeated but relatively
uninformative sequences. In gpt-5-mini, the power-law tendency becomes
stronger, while gpt-5 approaches the behavior observed in natural
language.  These results indicate that the relative fit of power-law
versus log-power models depends strongly on the model.

In contrast, for natural language texts, model preference varies
systematically with text length. Shorter texts tend to favor the
power-law model, suggesting continued introduction of new information,
whereas longer texts increasingly favor the log-power model. This
transition indicates that, as text length increases, structural reuse
through reference mechanisms becomes more prominent, leading to more
convex entropy growth and improved fit by the log-power model.
An extreme example is shown in \figref{fig:shakespeare}
(of \appref{app:shakespeare}), which plots
$\hat{H}_{\alpha}(m)$ for the complete works of Shakespeare
($n=5{,}442{,}126$).
Although this text contains a very long maximally repeated sequence,
the effective growth of information saturates at
moderate block lengths. This behavior strongly favors a log-power
function, and may even suggest an asymptotic form closer to a
log--log power law.

\subsection{Comparison with maximal repetition}
\label{sec:ablation}
\label{sec:eta}

\begin{figure}[t]
  \centering
  \includegraphics[width=\columnwidth]{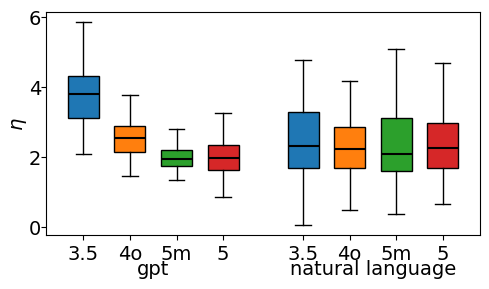}
\vspace*{-8mm}
\caption{Boxplots of the exponent $\eta$ governing the growth of
maximally repeated subsequences (baseline comparison). 
    \label{fig:boxeta}}
  \vspace*{-3mm}  
\end{figure}

For comparison with prior work, we analyze the growth of maximally
repeated subsequences.
A maximally repeated subsequence is defined as a subsequence
$x_i^{i+m_{\text{max}}}$ such that
\begin{equation}
m_{\text{max}} =
\max \left\{ \ell \;\middle|\; \exists\, i \neq j \text{ with }
x_i^{i+\ell} = x_j^{j+\ell} \right\}.
\end{equation}
It is known that
\begin{equation}
m_{\text{max}}(n) \propto (\log_2 n)^{\eta},
\end{equation}
where $\eta \to 1$ for i.i.d.\ sequences
\cite{OrnsteinWeiss1993}, whereas empirical studies of natural
language report $\eta > 1$ \cite{DebowskiMR}.

Figure~\ref{fig:boxeta} shows boxplots of the exponent
$\eta$ governing maximal repetition growth.
Overall, the qualitative trends observed for $\eta$ are consistent
with those obtained for the exponents $\beta$ and $\gamma$.
Earlier GPT models exhibit larger values of $\eta$, indicating the
presence of many long but relatively uninformative repetitions. This
tendency largely disappears in gpt-5, whose mean $\eta$ values become
comparable to those of natural language. In contrast, natural language
consistently shows a larger variance in $\eta$, while its mean values
remain stable across different text lengths, in line with the
observations for $\beta$ and $\gamma$.

Under maximal
repetition, gpt-5 behaves similarly to natural language apart from the
difference in variance, whereas the proposed approach
reveals systematically larger exponents for GPT-generated
texts, yielding a more pronounced distinction. Moreover, estimates
based on maximal repetition are numerically unstable due to their
reliance on extreme statistics, as discussed in \appref{sec:mrs},
which further limits their effectiveness as a diagnostic compared to
the distribution-based method proposed here.

Finally, our results also refine previous estimates of $\eta$ for
natural language. Prior work reported values around $\eta \approx 3$
\cite{DebowskiMR}, based in part on aggregated corpora such as the
complete works of Shakespeare, which exhibit substantial redundancy
due to corpus-level aggregation (see \appref{app:shakespeare}). In contrast, we find that for
individual texts, $\eta$ typically converges to values slightly above
$2$, as shown in \figref{fig:boxeta}, suggesting that earlier
estimates may reflect corpus composition. 

\section{Discussion}
\label{sec:discussion}

Our results refine the interpretation of entropy growth in
natural language by distinguishing between power-law and
logarithmic--power scaling. Although both appear sublinear over finite
ranges, they correspond to qualitatively different patterns of
information accumulation: power-law growth is consistent with a
continual increase in irreducible content, whereas
logarithmic--power growth reflects a slower expansion of
distinguishable structure.

This distinction can be interpreted through the lens of algorithmic
information theory. Sequences generated by simple mechanisms, such as
short programs that repeatedly reuse the same rules, admit
descriptions whose Kolmogorov complexity grows much more slowly than
sequence length \cite{Kolmogorov1965,LiVitanyi2008}. By analogy,
increases in description length may be dominated by specifying
references to previously defined structures rather than encoding new
content.

Applied to language, a log-power fit is consistent with strong
structural reuse, where texts are generated through recombination and
re-indexing of shared linguistic resources, yielding sublinear growth
in informational complexity.
This view remains compatible with
Zipfian frequency distributions and unbounded vocabulary growth
\cite{Zipf1949,Herdan1964,Heaps1978}, since lexical innovation can
coexist with slower growth in syntactic and discourse-level
structure.

Within this framework, the differences observed between natural
language and GPT-generated text suggest a weaker manifestation of this
structure-as-reference organization in current models. Although model
outputs exhibit repetition and long-range dependencies, the estimated
exponents and their dependence on R\'enyi order indicate a greater
reliance on locally generated structure, possibly due to the
next-token prediction objective and finite context utilization.

These results provide empirical support for viewing natural language
as operating near a regime dominated by structural reuse. Further
work is needed to clarify how training objectives and architectures
shape repetition-based entropy growth.

\section{Conclusion}

We introduced a repetition-based framework for analyzing long-range
structure in natural language and large language model outputs. By
relating repeated subsequences to higher-order R\'enyi entropies and
accounting for finite-size effects, we showed that entropy growth is
often better characterized by logarithmic--power than by
power-law models.

Across datasets, natural language exhibits stable and strongly
sublinear growth patterns, whereas GPT-generated texts show systematic
and statistically significant shifts in exponents with model
size. These results reveal persistent differences in long-range
statistical organization that are not captured by standard
task-based evaluations.

Overall, our findings establish repeated-subsequence entropy as a
quantitative diagnostic for long-range structure, providing a simple
way in finite-length data to assess how closely model-generated text
matches the statistical patterns of natural language.

\section{Limitations}

The limitations discussed below fall into three categories:
scope, methodology, and interpretation.

\paragraph{Scope limitations.}
A first limitation concerns the scope of the data.
The present analysis focuses on structural reuse within individual
texts, whereas classical studies of Shannon entropy ($H_1$) typically
analyze large heterogeneous corpora aggregated across multiple
authors, topics, and styles, rather than single coherent documents.
As a result, our approach captures intra-text organization but does
not directly address cross-text variability or population-level
statistics. Extending the framework to larger-scale datasets that
enable analysis of both within-text and across-text structure remains
an important direction for future work.

A second limitation is that our experiments focus on GPT-family
models as representative large language models. Although these
models span multiple generations and sizes, the results do not
necessarily generalize to all architectures or training paradigms.
Models with different objectives, architectures, or external memory
components may exhibit different entropy-growth behavior.

A third limitation is that our analysis is conducted at the
alphabetic character level.
This choice avoids tokenization-specific artifacts and enables a
clean information-theoretic treatment, but does not directly operate
on higher-level linguistic units such as words, morphemes, or
syntactic constructions.

\paragraph{Methodological limitations.}
A fourth limitation is that our analysis does not directly resolve
the presence of an extensive linear component in entropy growth.
Because we focus on the growth behavior of $H_{\alpha}(m)$ rather
than on the ratio $H_{\alpha}(m)/m$, we do not obtain a direct
estimate of the entropy rate $h_{\alpha}$ or determine whether a
linear term emerges at larger scales.

A fifth limitation concerns the range of block lengths considered.
Our conclusions are based on entropy-growth behavior over
empirically accessible values of $m$. Although we account for
finite-size effects, longer texts or alternative estimation methods
may reveal additional structure beyond the ranges considered here.

\paragraph{Interpretative limitations.}
A sixth limitation is that the proposed approach is not designed to
measure task performance or downstream capabilities. Instead, it
provides a complementary structural signal. Understanding how these
measures relate to functional performance remains an open question.

Finally, while we observe systematic and statistically significant
differences between natural language and LLM-generated text, our
analysis is descriptive and does not identify the mechanisms
responsible for these differences. Establishing causal links between
entropy-growth patterns and model properties remains an important
direction for future work.

\section*{Acknowledgements}

This work was supported by JST CREST, Japan, Grant Number JPMJCR2114.

AI assistants were used during the development and writing of this
work. All technical content, analyses, and conclusions are the sole
responsibility of the author.

\bibliography{repeats}


\renewcommand{\appendixname}{Supplementary}
\section*{\Large Supplementary Material}
\appendix

\section{Repeated Sequences for a Bernoulli Process ($p=0.5$)}
\label{app:binaryDup}

\begin{figure}[h]
  \centering
  \includegraphics[width=\columnwidth]{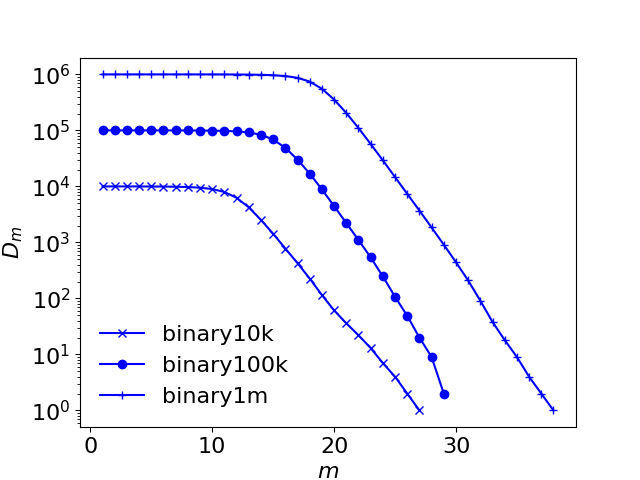}
  \vspace*{-5mm}
  \caption{Repeated subsequences $D_m$ for a Bernoulli process with
    $p=0.5$ ($n=10$k, $100$k, and $1$m).
    \label{fig:binaryDup}}
\end{figure}

\figref{fig:binaryDup} shows the counts of repeated subsequences $D_m$
(vertical axis, logarithmic scale) as a function of block length $m$
(horizontal axis) for Bernoulli processes of length $n=10$k, $100$k,
and $1$m. Longer sequences appear higher in the plot, reflecting the
larger number of available blocks.

As discussed in the main text (\secref{sec:binaryDup}), shuffled
natural language texts exhibit the same qualitative behavior as the
Bernoulli process. For small $m$, essentially all $|\mathcal{A}|^m$
possible blocks appear in the sequence. As $m$ increases, the number
of repeated blocks decreases sharply. This transition corresponds to
the well-known \emph{birthday limit}
\cite{MotwaniRaghavan1995,Knuth1997TAOCP2}.

\section{$h_{\alpha}=1$ for a Bernoulli Process with $p=0.5$}
\label{app:halpha1}

For a Bernoulli process, the following derivation applies.
Let the probability of symbol $1$ be $p$.
Then the probability of a word $w$ of length $m$ containing $k$ ones is
given by
\[
  p_{w} = p^{k}(1-p)^{m-k}.
\]
Therefore,
\begin{eqnarray*}
  \sum_{|w|=m} p_{w}^{\alpha}
  &=& \sum_{k=0}^{m} \binom{m}{k} (p^{\alpha})^{k}
    \bigl((1-p)^{\alpha}\bigr)^{m-k} \\
  &=& \left(p^{\alpha} + (1-p)^{\alpha}\right)^{m}.
\end{eqnarray*}
Consequently,
\begin{eqnarray*}
  H_{\alpha}(m)
  &=& \frac{1}{1-\alpha}
      \log_2 \left( \sum_{|w|=m} p_{w}^{\alpha} \right) \\
  &=& \frac{1}{1-\alpha}
      \log_2 \left( p^{\alpha} + (1-p)^{\alpha} \right)^{m}.
\end{eqnarray*}
In particular, when $p=0.5$, the R\'enyi entropy grows linearly with $m$
with slope~$1$, independently of the value of $\alpha$.

\section{Rationale for Probability Estimation of $p_w$}
\label{app:p}

Throughout this paper, we estimate the block probability $p_w$ by the
empirical frequency
\[
\hat p_w = \frac{\#w}{T_m},
\]
where $\#w$ denotes the number of occurrences of block $w$ among the
$T_m = n - m + 1$ length-$m$ blocks extracted from a sequence of length
$n$. This choice corresponds to defining $p_w$ as the probability that
block $w$ is observed when a start position is sampled uniformly at
random from the $T_m$ available positions.

Under stationarity and ergodicity, this estimator is consistent for
the true block probability $\Pr(X_1^m = w)$.
More generally, for non-stationary texts, $\hat p_w$ estimates the
position-averaged block distribution, which reflects the empirical
statistics of the finite sequence under uniform sampling of block
positions. This operational definition is well aligned with our
finite-length analysis of repeated subsequences.

In contrast, normalizing by the number of possible block types $S_m$
would not yield a meaningful probability distribution over observed
blocks, since $S_m$ counts types rather than sampling trials.
Moreover, our finite-size corrections and occupancy-based derivations,
including the definition of $\lambda_m = T_m / S_m$, are naturally
formulated in terms of $T_m$ as the number of block occurrences.

Furthermore, in this work, $\hat{H}_{\alpha}(m)$ is computed using only
blocks with $\#w \geq \alpha$. The first reason is conceptual: the
R\'enyi order $\alpha$ directly corresponds to the degree of repetition
being emphasized, and blocks occurring fewer than $\alpha$ times do
not meaningfully contribute to this regime. The second reason is
empirical: when rarer blocks with $\#w < \alpha$ are included, the
empirical estimates for a Bernoulli process systematically deviate
from the theoretical predictions and fail to converge to the expected
behavior.

\section{Derivation of $\Delta_2$}
\label{app:delta}

To derive formula~(\ref{eq:base}) in the main text, we consider the
case $\alpha=2$.
Assume that words of length $m$ occur independently according to a
Poisson distribution with mean $\lambda_m = T_m / S_m$.
Among the $S_m$ possible word types, $K_m$ types are observed in the
sequence.
Let $C_i$ denote the number of occurrences of the $i$-th word type.
For $\alpha=2$, we have
\begin{equation*}
  \sum_{i=1}^{K_m} p_i^2
  = \sum_{i=1}^{S_m} \left( \frac{C_i}{T_m} \right)^2
  = \frac{1}{T_m^2} \sum_{i=1}^{S_m} C_i^2 .
\end{equation*}
Since unobserved word types satisfy $C_i = 0$, the summation may be
taken over all $S_m$ types.
Because $C_i \sim \mathrm{Poisson}(\lambda_m)$, we obtain
\begin{equation*}
  E[C_i^2] = \mathrm{Var}(C_i) + (E[C_i])^2
  = \lambda_m + \lambda_m^2.
\end{equation*}
Assuming independence, it follows that
\begin{equation*}
  E\!\left[ \sum_{i=1}^{S_m} C_i^2 \right]
  = \sum_{i=1}^{S_m} E[C_i^2]
  = S_m (\lambda_m^2 + \lambda_m).
\end{equation*}
Therefore,
\begin{equation*}
  E\!\left[ \sum_{i=1}^{K_m} p_i^2 \right]
  = \frac{S_m (\lambda_m^2 + \lambda_m)}{T_m^2}.
\end{equation*}
Substituting $T_m = S_m \lambda_m$, we obtain
\begin{equation*}
  E\!\left[ \sum_{i=1}^{K_m} p_i^2 \right]
  = \frac{1}{S_m} \left( 1 + \frac{1}{\lambda_m} \right).
\end{equation*}

To take logarithms, we use the approximation
\begin{equation*}
  E[-\log X] \approx -\log E[X],
\end{equation*}
which yields
\begin{eqnarray*}
  H_2(m)
  &=& -\log \left[ \frac{1}{S_m}
      \left( 1 + \frac{1}{\lambda_m} \right) \right] \\
  &=& \log S_m - \log \left( 1 + \frac{1}{\lambda_m} \right).
\end{eqnarray*}
This yields the desired form, with correction term
$\Delta_2 = \log \left( 1 + \frac{1}{\lambda_m} \right)$.
As noted in the main text, these correction terms can be expressed as
weighted logarithms of Touchard polynomials \cite{Touchard1956}.
Similarly, for $\alpha=3$ and $\alpha=4$, the corresponding correction
terms $\Delta_3$ and $\Delta_4$ are given by as
\begin{eqnarray*}
  \Delta_3 &=& \frac{1}{2}
  \log\!\left( 1 + \frac{3}{\lambda_m} + \frac{1}{\lambda_m^2} \right),
 \\
  \Delta_4 &=& \frac{1}{3}
  \log\!\left( 1 + \frac{6}{\lambda_m}
  + \frac{7}{\lambda_m^2}
  + \frac{1}{\lambda_m^3} \right).
\end{eqnarray*}

\section{Relation between $\lambda_m$ and $D_m$}
\label{app:lambda}

It can be shown that the relationship between
$\lambda_m = T_m / S_m$ and $D_m$ is given by
\begin{equation}
  E[D_m] = T_m f(\lambda_m),
\end{equation}
where
\begin{equation}
  f(\lambda) \equiv \left(1 - \frac{1 - e^{-\lambda}}{\lambda}\right).
\end{equation}

When $p_{w}$ is uniformly distributed (i.e., $p_{w} = 1 / S_m$), we have
\begin{equation}
  \sum_{|w|=m} e^{-T_m p_{w}} = S_m e^{-T_m / S_m},
\end{equation}
and hence, from formula~(\ref{eq:dm}) in the main text,
\begin{eqnarray*}
  E[D_m] &\approx& T_m - S_m (1 - e^{-T_m / S_m}) \\
         &=& T_m \left(1 - \frac{1 - e^{-T_m / S_m}}{T_m / S_m}\right) \\
         &=& T_m \left(1 - \frac{1 - e^{-\lambda_m}}{\lambda_m}\right).
\end{eqnarray*}

For a general distribution $p_{w}$, we use the expansion of
formula~(\ref{eq:expand}) in the main text, whose first-order term
cancels out. This implies that the probability of repetition is
determined by the second moment.
Following the standard technique of \emph{effective uniformization},
we introduce a uniform distribution with the same second moment
$q_m = \sum_{|w|=m} p_w^2$.
Let $S_{\text{eff}} = 1 / q_m$, and define
\begin{equation}
  \tilde{p}_w \equiv \frac{1}{S_{\text{eff}}}.
\end{equation}
Then,
\begin{eqnarray}
  \sum_{|w|=m} \tilde{p}_w^2 &=& S_{\text{eff}}
  \left(\frac{1}{S_{\text{eff}}}\right)^2 \\
  &=& \frac{1}{S_{\text{eff}}} \\
  &=& q_m.
\end{eqnarray}

Letting $\lambda_m \equiv T_m q_m$, where $q_m$ is the second moment of
the original distribution, we can apply the same argument as in the
uniform case to obtain
\begin{eqnarray}
  E[D_m] &\approx& T_m - S_{\text{eff}} (1 - e^{-T_m / S_{\text{eff}}}) \\
         &=& T_m f(\lambda_m),
\end{eqnarray}
which is independent of the specific form of $\tilde{p}_w$.

\section{GPT Text Generation}
\label{app:gpt}

GPT models, particularly earlier versions, impose practical limits on
the maximum length of a single response. To generate sufficiently
long texts for our analysis, we adopt a strategy in which each model
is instructed to produce a single coherent story in multiple parts.

Specifically, for each GPT model, we provide the following prompt:
\begin{quote}
You are a genius storyteller. I want you to generate a story longer
than $numWords$ tokens in $num$ parts. Please tell the story part by
part.
\end{quote}

The target length $numWords$ is set to $200{,}000$ tokens, reflecting
the typical scale of long human-written narratives (on the order of
$100{,}000$ tokens). The number of parts $num$ is set to $20$, so that
each generated text is produced incrementally while maintaining
global narrative consistency.

\section{Domain Effects and Robustness}
\label{app:domain}

\begin{table}[t]
\centering
  \caption{Datasets generated for Essay and Scientific Articles}
  \label{tab:datasetdomain}
  \vspace*{-2mm}
  \footnotesize
  \begin{tabular}{|l|l|r|}
    \hline
  dataset & number  & length (chars) \\
    \hline
    \multicolumn{3}{|c|}{Essay} \\
    \hline
gpt-3.5turbo & 30 & 36489.23$\pm$2368.41 \\
gpt-4o-mini  & 30 & 28196.33$\pm$7772.60 \\
gpt-5-mini   & 30 & 157692.33$\pm$20627.09 \\
gpt-5        & 30 & 313300.67$\pm$11774.73 \\
    \hline
    \multicolumn{3}{|c|}{Scientific Article} \\
    \hline
gpt-3.5turbo & 30 & 39002.53$\pm$5422.58 \\
gpt-4o-mini  & 30 & 60039.40$\pm$8858.72 \\
gpt-5-mini   & 30 & 193521.70$\pm$25617.89 \\
gpt-5        & 30 & 370674.73$\pm$12729.61 \\
    \hline
  \end{tabular}
\end{table}

\begin{figure}[t]
  \centering
  \includegraphics[width=\columnwidth]{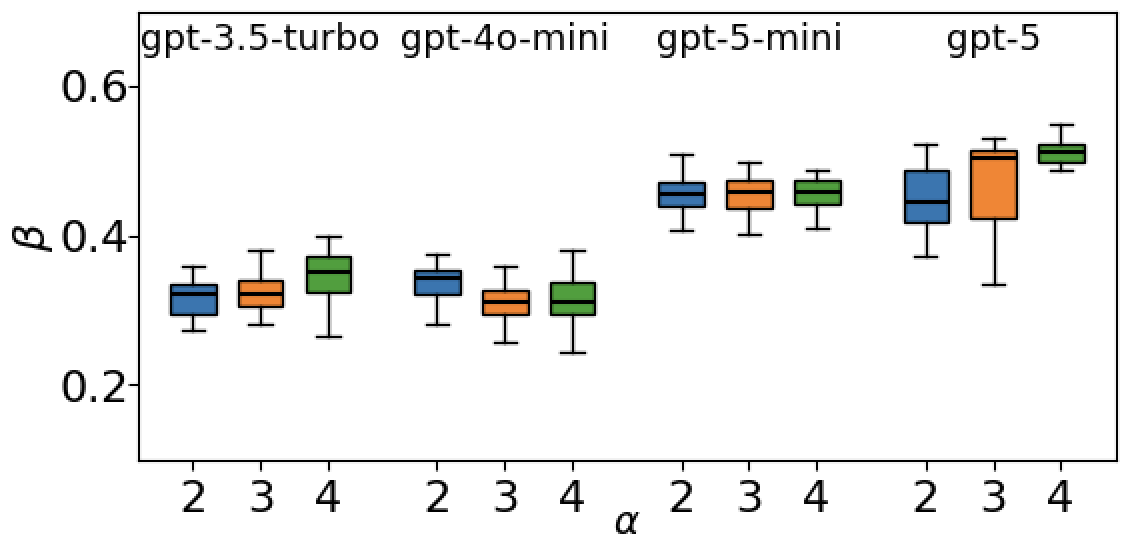}
  \includegraphics[width=\columnwidth]{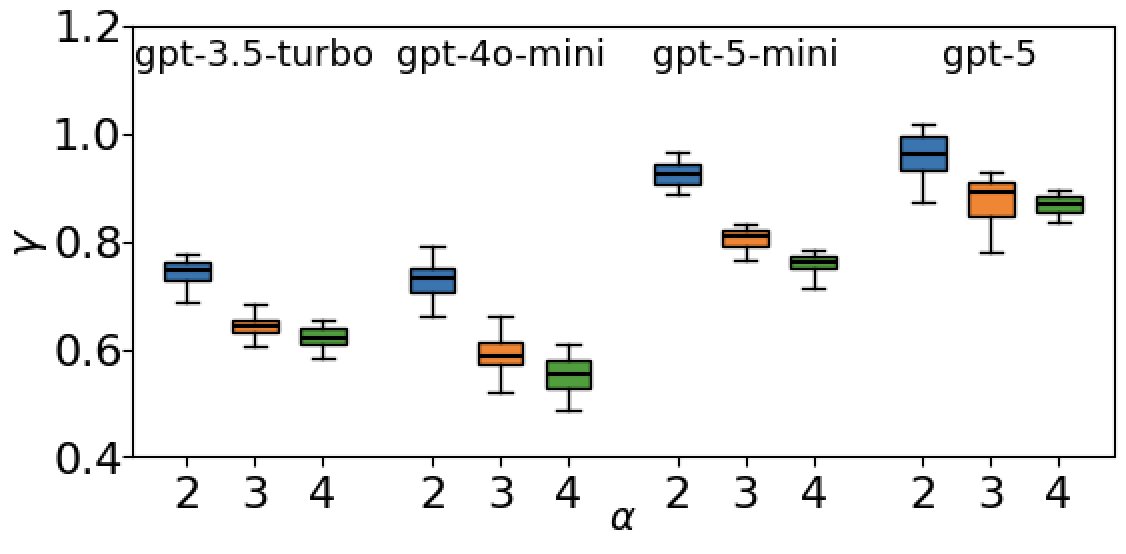}
  \vspace*{-8mm}
  \caption{Boxplots of $\beta$ and $\gamma$ for GPT generated Essay dataset.
    \label{fig:boxbeta_essay}}
\end{figure}
\begin{figure}[t]
  \centering
  \includegraphics[width=\columnwidth]{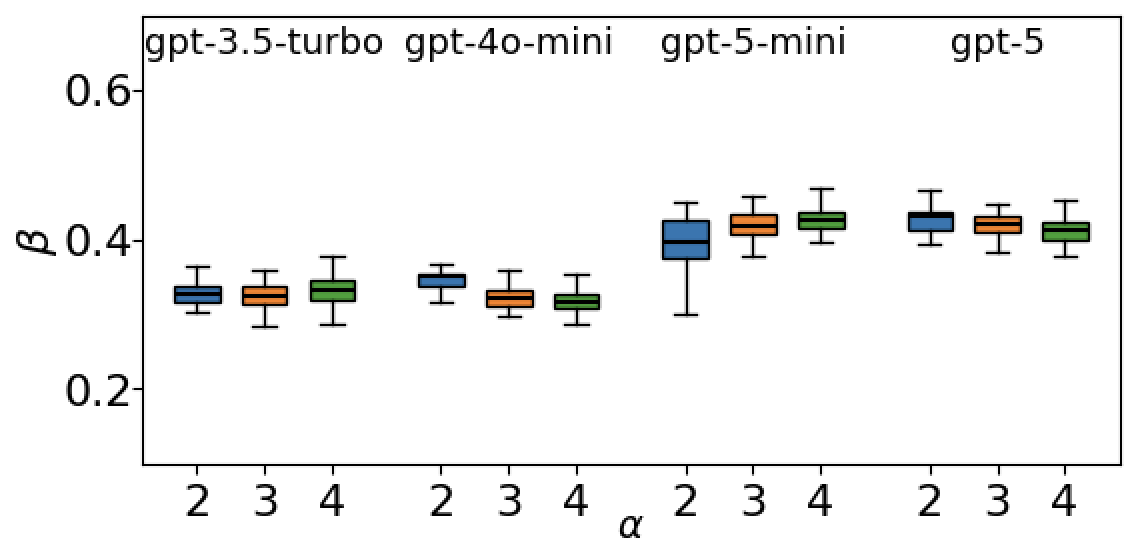}
  \includegraphics[width=\columnwidth]{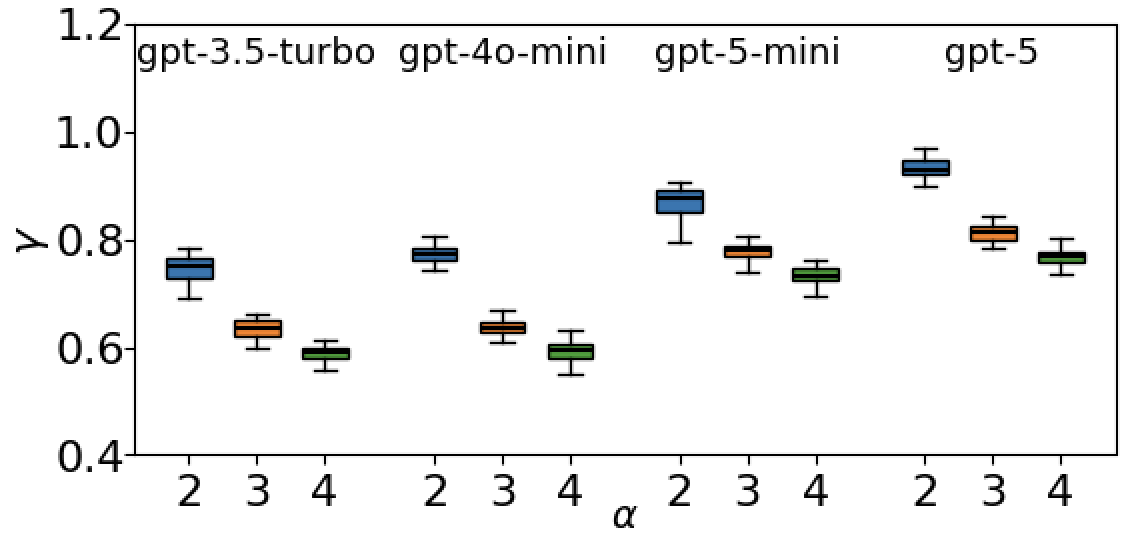}
  \vspace*{-8mm}
  \caption{Boxplots of $\beta$ and $\gamma$ for GPT generated Scientific Article dataset.
    \label{fig:boxbeta_science}}
\end{figure}

To examine domain dependence, we compare results across essays and
scientific-style texts.

For natural language, obtaining texts of comparable length and
coherence to novels in these domains is challenging. As noted in the
main text, it is difficult to collect sufficiently long and
structurally consistent essays or scientific articles, and the wide
variation in topics further limits assembling well-matched data.
Extending the analysis to large-scale corpora spanning multiple
authors and domains therefore remains an important direction for
future work.

In contrast, GPT-generated texts allow controlled generation of
alternative domains. Using the same prompts as in the previous
section, we generated 30 essays and 30 scientific-style texts by
replacing ``story''. The resulting datasets are summarized in
\tabref{tab:datasetdomain}. 
These texts are approximately half as long as the generated stories,
consistent with the tendency of human-written essays and scientific
texts to be shorter than novels.

For these datasets, we constructed boxplots of the estimated 
exponents, analogous to those in \secref{sec:betagamma}. The results
for essays are broadly consistent with those for novels in terms of
distributions and model-dependent trends, although model preference
favors power-law in a large majority of cases.

In contrast, scientific-style texts exhibit a quantitatively different
pattern while preserving the same qualitative trends.  The estimated
values of $\beta$ and $\gamma$ are systematically smaller, and model
preference favors power-law model over log-power model in most
cases. At the same time, the dependence on model size remains
consistent with the main analysis: both $\beta$ and $\gamma$ increase
with model size, and their dependence on $\alpha$ follows the same
pattern, with $\beta$ increasing and $\gamma$ decreasing as $\alpha$
grows.

These observations indicate that the overall structural behavior is
shared across domains, but that its strength differs.
In particular, the weaker log-power preference and smaller exponent
values suggest that referential reuse is more limited in
GPT-generated scientific-style texts.
Overall, the results highlight that domain primarily affects the
strength of entropy growth, while the overall trends remain similar.

\begin{table*}[h]
\centering
\caption{Mean and standard deviation of the estimated exponent
$\beta$ (mean $\pm$ std), together with the coefficient of determination
$R^2$, for $\alpha=2,3,4$ across all datasets.\label{tab:beta}}
\vspace*{-3mm}
\footnotesize
\begin{tabular}{lccc}
\hline
& $\alpha=2$ & $\alpha=3$ & $\alpha=4$ \\
\hline
    \multicolumn{4}{c}{GPT-generated text} \\
\hline
gpt-3.5
& $0.347\pm0.032\;(0.956\pm0.014)$
& $0.344\pm0.032\;(0.890\pm0.027)$
& $0.389\pm0.053\;(0.878\pm0.051)$ \\
gpt-4o-mini
& $0.424\pm0.026\;(0.976\pm0.010)$
& $0.424\pm0.030\;(0.942\pm0.022)$
& $0.446\pm0.034\;(0.932\pm0.029)$ \\
gpt-5-mini
& $0.505\pm0.045\;(0.977\pm0.037)$
& $0.518\pm0.017\;(0.976\pm0.010)$
& $0.524\pm0.014\;(0.970\pm0.008)$ \\
gpt-5
& $0.485\pm0.051\;(0.960\pm0.042)$
& $0.520\pm0.031\;(0.967\pm0.027)$
& $0.533\pm0.030\;(0.964\pm0.027)$ \\
\hline
\multicolumn{4}{c}{Natural language text} \\
\hline
nl-3.5
& $0.414\pm0.103\;(0.952\pm0.054)$
& $0.427\pm0.106\;(0.933\pm0.068)$
& $0.439\pm0.108\;(0.920\pm0.078)$ \\
nl-4o
& $0.406\pm0.088\;(0.941\pm0.063)$
& $0.418\pm0.091\;(0.920\pm0.075)$
& $0.425\pm0.101\;(0.910\pm0.079)$ \\
nl-5-mini
& $0.396\pm0.094\;(0.933\pm0.056)$
& $0.419\pm0.103\;(0.923\pm0.078)$
& $0.431\pm0.105\;(0.921\pm0.073)$ \\
nl-5
& $0.371\pm0.108\;(0.913\pm0.072)$
& $0.415\pm0.096\;(0.919\pm0.072)$
& $0.429\pm0.100\;(0.917\pm0.071)$ \\
\hline
\end{tabular}
\end{table*}

\begin{table*}[h]
\centering
\caption{Mean and standard deviation of the estimated exponent
$\gamma$ (mean $\pm$ std), together with the coefficient of determination
$R^2$, for $\alpha=2,3,4$ across all datasets.\label{tab:gamma}}
\vspace*{-3mm}
\footnotesize
\begin{tabular}{lccc}
\hline
 & $\alpha=2$ & $\alpha=3$ & $\alpha=4$ \\
\hline
\multicolumn{4}{c}{GPT-generated text} \\
\hline
gpt-3.5
& $0.773\pm0.031\;(0.953\pm0.014)$
& $0.690\pm0.035\;(0.965\pm0.007)$
& $0.703\pm0.050\;(0.950\pm0.010)$ \\
gpt-4o-mini
& $0.894\pm0.023\;(0.963\pm0.013)$
& $0.807\pm0.026\;(0.975\pm0.005)$
& $0.795\pm0.026\;(0.966\pm0.005)$ \\
gpt-5-mini
& $1.031\pm0.041\;(0.965\pm0.013)$
& $0.942\pm0.011\;(0.970\pm0.007)$
& $0.910\pm0.011\;(0.966\pm0.006)$ \\
gpt-5
& $1.043\pm0.048\;(0.976\pm0.012)$
& $0.971\pm0.023\;(0.975\pm0.006)$
& $0.944\pm0.021\;(0.972\pm0.005)$ \\
\hline
\multicolumn{4}{c}{Natural language text} \\
\hline
nl-3.5
& $0.804\pm0.120\;(0.937\pm0.045)$
& $0.705\pm0.108\;(0.929\pm0.055)$
& $0.669\pm0.111\;(0.919\pm0.051)$ \\
nl-4o
& $0.848\pm0.108\;(0.941\pm0.068)$
& $0.755\pm0.107\;(0.929\pm0.096)$
& $0.716\pm0.117\;(0.917\pm0.096)$ \\
nl-5-mini
& $0.892\pm0.112\;(0.961\pm0.031)$
& $0.818\pm0.119\;(0.952\pm0.038)$
& $0.793\pm0.114\;(0.946\pm0.038)$ \\
nl-5
& $0.882\pm0.142\;(0.944\pm0.064)$
& $0.844\pm0.122\;(0.945\pm0.067)$
& $0.818\pm0.122\;(0.937\pm0.076)$ \\
\hline
\end{tabular}
\end{table*}

\section{Actual Values Used to Produce Figures~\ref{fig:boxbeta}--\ref{fig:boxeta}}
\label{app:vals}

Tables~\ref{tab:beta}--\ref{tab:eta_rsd} report the numerical values
used to generate Figures~\ref{fig:boxbeta}--\ref{fig:boxeta},
respectively.  Interpretation and discussion of these results are
provided in the main text.

\begin{table}[ht]
\centering
\caption{Number of texts for which the log-power model provides a better
fit than the power-law model (LP $>$ P).\label{tab:lpp}}
\vspace*{-3mm}
\footnotesize
\begin{tabular}{lccc}
\hline
& $\alpha=2$ & $\alpha=3$ & $\alpha=4$ \\
\hline
\multicolumn{4}{c}{GPT-generated text} \\
\hline
gpt-3.5 & 40 & 99 & 97 \\
gpt-4o-mini & 24 & 93 & 90 \\
gpt-5-mini & 19 & 29 & 32 \\
gpt-5 & 60 & 52 & 58 \\
\hline
\multicolumn{4}{c}{Natural language text} \\
\hline
nl-3.5 & 33 & 40 & 42 \\
nl-4o  & 46 & 57 & 54 \\
nl-5-mini & 66 & 58 & 54 \\
nl-5  & 66 & 68 & 61 \\
\hline
\end{tabular}
\end{table}
\begin{table}[t]
\centering
\caption{Exponent $\eta$ of maximal repetition growth.  Values are
  reported as mean $\pm$ standard deviation together with the
  coefficient of determination $R^2$.  The column ``fail'' indicates
  the number of texts for which the estimation did not converge. Each
  category contains 100 instances.  \label{tab:eta_rsd}}
\footnotesize
\begin{tabular}{lccc}
\hline
Dataset & $\eta$ (mean $\pm$ std) & $R^2$ & fail \\
\hline
\multicolumn{4}{c}{GPT-generated text} \\
\hline
gpt-3.5-turbo & $3.815 \pm 0.929$ & $0.812 \pm 0.119$ & 0 \\
gpt-4o-mini  & $2.608 \pm 0.603$ & $0.862 \pm 0.085$ & 0 \\
gpt-5-mini   & $2.091 \pm 0.616$ & $0.829 \pm 0.113$ & 0 \\
gpt-5        & $2.085 \pm 0.646$ & $0.810 \pm 0.138$ & 0 \\
\hline
\multicolumn{4}{c}{Natural language text} \\
\hline
nl-3.5       & $2.855 \pm 2.172$ & $0.686 \pm 0.235$ & 1 \\
nl-4o        & $2.732 \pm 2.075$ & $0.709 \pm 0.225$ & 6 \\
nl-5-mini    & $2.598 \pm 1.788$ & $0.720 \pm 0.226$ & 1 \\
nl-5         & $2.454 \pm 1.263$ & $0.715 \pm 0.229$ & 7 \\
\hline
\end{tabular}
\end{table}

\begin{figure*}[h]
  \centering
  \includegraphics[width=0.24\textwidth]{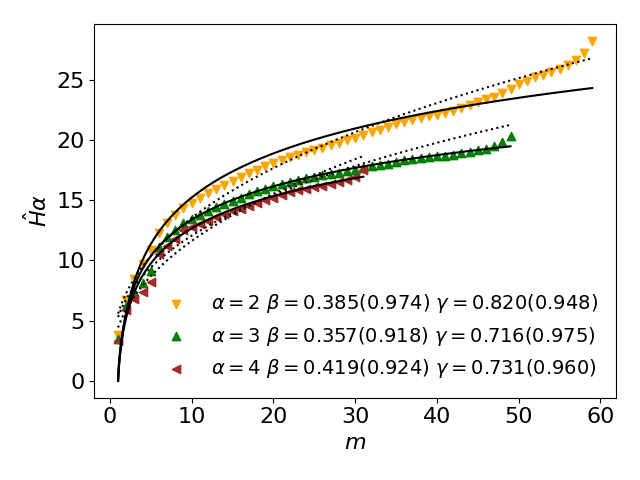}
  \includegraphics[width=0.24\textwidth]{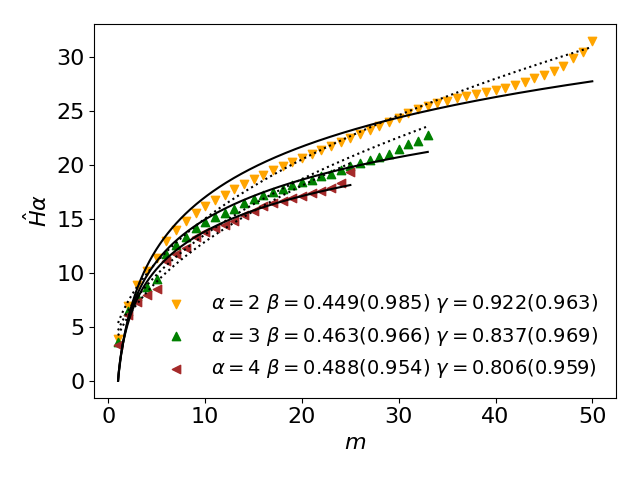}
  \includegraphics[width=0.24\textwidth]{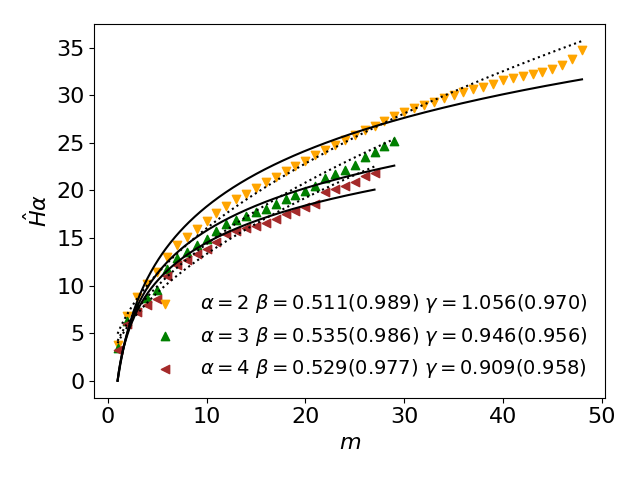}  
  \includegraphics[width=0.24\textwidth]{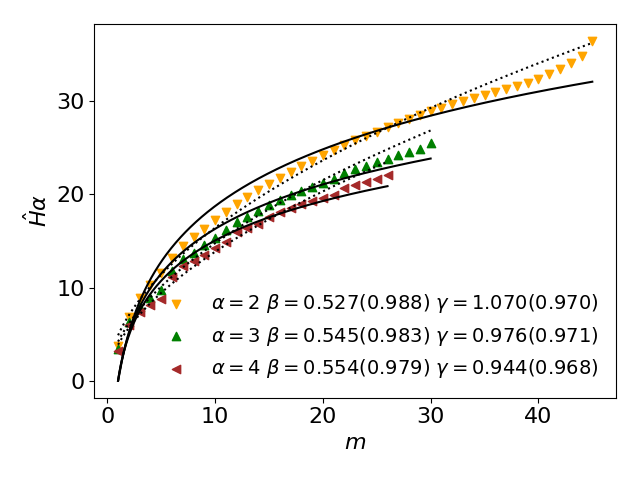}
  \vspace*{-5mm}
  \caption{Empirical higher-order R\'enyi entropy $\hat{H}_{\alpha}(m)$
    for a representative sample generated by gpt-3.5-turbo
    4o-mini, 5-mini and 5.
    \label{fig:h2_gpthalpha}}
\end{figure*}

\section{$\hat{H}_{\alpha}(m)$ for Samples of GPT-Generated Texts}
\label{app:gpthalpha}

\figref{fig:h2_gpthalpha} shows $\hat{H}_{\alpha}(m)$ for one
representative sample from each of the GPT models considered
(gpt-3.5-turbo, gpt-4o-mini, gpt-5-mini, and gpt-5).

For gpt-3.5-turbo and gpt-4o-mini, the empirical curves exhibit a
clear preference for the log-power functional form at higher Rényi
orders ($\alpha=3,4$), indicating increasingly convex entropy growth.
In contrast, for $\alpha=2$, GPT-generated texts typically favor the
power-law model across all versions, suggesting that these models
continue to introduce new block types even at large text lengths.


   


\section{$H_{\alpha}(m)$ of Shakespeare}
\label{app:shakespeare}

\begin{figure}[ht]
  \centering
  \includegraphics[width=\columnwidth]{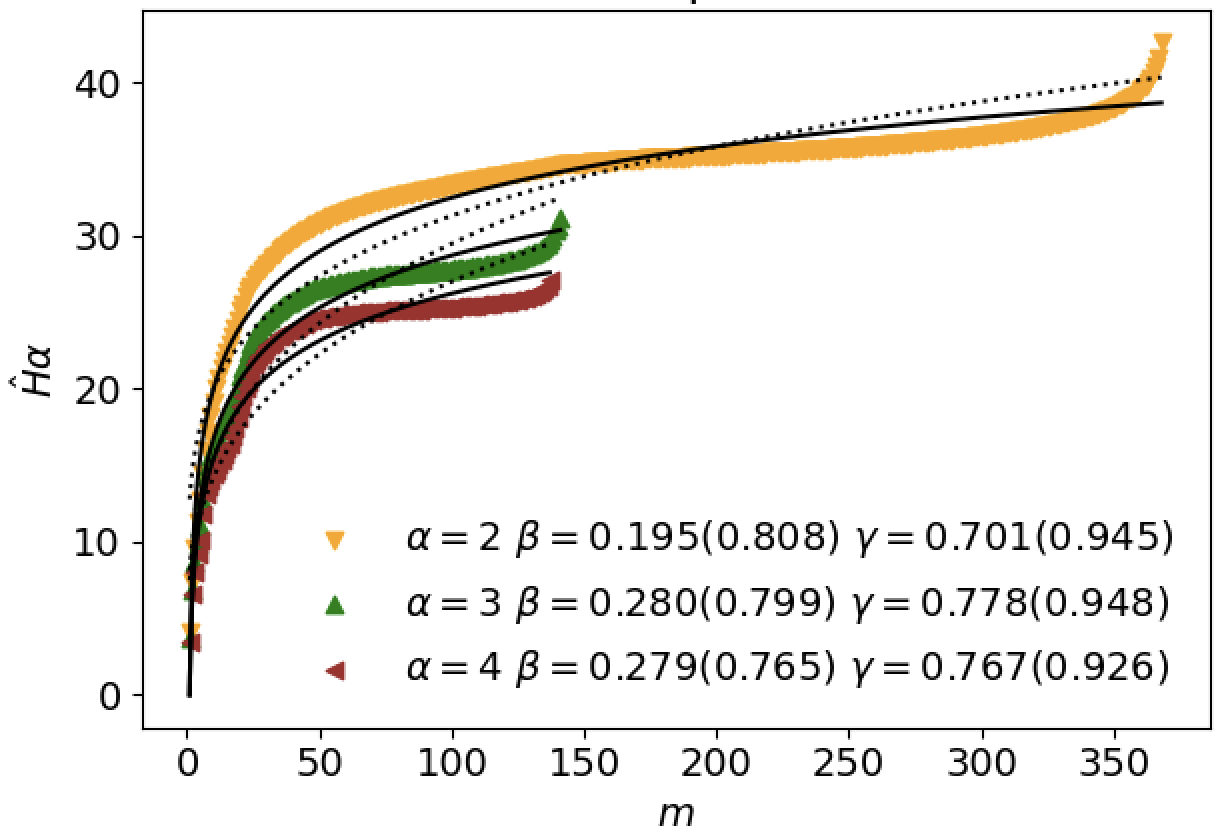}
  \vspace*{-6mm}
  \caption{Empirical higher-order R\'enyi entropy
    $\hat{H}_{\alpha}(m)$ for $\alpha=2,3,4$ computed on the complete
    works of Shakespeare.
    \label{fig:shakespeare}}
\end{figure}

\figref{fig:shakespeare} shows $\hat{H}_{\alpha}(m)$ for the complete
works of Shakespeare. Although repeated subsequences are observed up
to block lengths of approximately $m \approx 350$, the effective
growth of entropy continues only up to around $m \approx 50$, beyond
which the increase in $\hat{H}_{\alpha}(m)$ becomes very slow. This
figure therefore illustrates an extreme case of long-range
repetition. In contrast, for typical individual texts,
$\hat{H}_{\alpha}(m)$ more closely resembles the behavior shown in
\figref{fig:h2-hfe} (bottom), where effective entropy growth decreases
gradually and smoothly as $m$ increases.

\section{Maximal Repetition: Definition, Examples, and Limitations}
\label{sec:mrs}



\begin{figure}[h]
  \centering
  \includegraphics[width=\columnwidth]{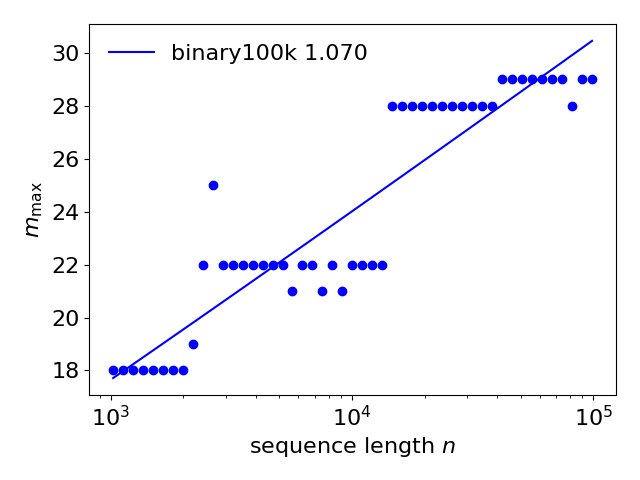}
 \vspace*{-5mm}
  \caption{Maximal repetition length for a Bernoulli process.}
  \label{fig:mrs-binary}
\end{figure}

\begin{figure}[t]
  \centering
  \includegraphics[width=0.9\columnwidth]{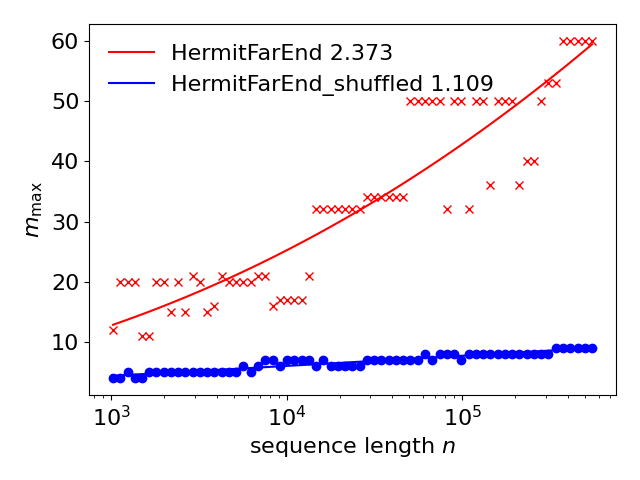}
 \vspace*{-5mm}
  \caption{Growth of the maximally repeated subsequence length for a
    natural language text (red) and its shuffled counterpart (blue).}
  \label{fig:mrs}
 \vspace*{-3mm}
\end{figure}

\figref{fig:mrs-binary} shows the behavior for a Bernoulli process
with $p=0.5$, where $\eta \approx 1$, consistent with theory.
Figure~\ref{fig:mrs} shows results for \myex. 
The shuffled version yields $\eta \approx 1$, whereas
the original text gives $\eta \approx 2.4$, indicating stronger
repetition.

The growth curves display a stepwise structure, reflecting the fact
that only a very small number of subsequences attain the maximal
length. As a result, the estimates are numerically unstable and
converge slowly, and for a non-negligible fraction of texts, $\eta$
cannot be reliably estimated.

To reduce noise, the curves in \figref{fig:mrs} were computed using
logarithmically spaced sampling points, where each point represents
the median over $k$ samples (with $k=5$). Despite this smoothing,
the estimates remain unstable, as illustrated by the shuffled text,
for which $\eta = 1.1$.

These limitations motivate the use of distribution-based measures,
which capture repetition across scales rather than relying on extreme
statistics.

\end{document}